\documentclass[16pt,...]{article}
\usepackage[papersize={216mm,330mm},tmargin=15mm,bmargin=15mm,lmargin=15mm,rmargin=15mm]{geometry}
\usepackage{physics}
\usepackage[utf8]{inputenc}
\usepackage{amsmath,amssymb,amsthm,amsfonts,bm}
\usepackage{tabularx}
\usepackage{graphicx}
\usepackage{rotating}
\usepackage{lscape}
\usepackage{float}
\usepackage{booktabs}
\usepackage{subcaption}
\usepackage{relsize}
\usepackage{color}
\usepackage{hyperref}
\usepackage[final]{pdfpages}
\usepackage{multirow}
\usepackage[affil-it]{authblk} 
\usepackage{etoolbox}
\usepackage{lmodern}
\usepackage{cite}
\usepackage[symbol]{footmisc}

\usepackage{lipsum}

\makeatletter
\renewcommand*{\@fnsymbol}[1]{\ensuremath{\ifcase#1\or \dagger\or \dagger\or *\or
    \mathsection\or \mathparagraph\or \|\or **\or \dagger\dagger
    \or \ddagger\ddagger \else\@ctrerr\fi}}
\makeatother
\title{Modelling and optimization of nanovector synthesis for applications in drug delivery systems}

\author[1]{Felipe J. Villaseñor-Cavazos\thanks{Felipe Villaseñor-Cavazos and Daniel Torres-Valladares contributed equally to this work.}}
\author[1]{Daniel Torres-Valladares$^{\dagger}$}
\author[1,2]{Omar Lozano \thanks{Correspondence: omar.lozano@tec.mx}} 

\affil[1]{Cátedra de Cardiología y Medicina Vascular, Escuela de Medicina y Ciencias de la Salud, Tecnologico de Monterrey, Av. Eugenio Garza Sada
2501 Sur, Monterrey 64849, Mexico}
 
 \affil[2]{Centro de Investigación Biomédica, Hospital Zambrano-Hellion, Tecnologico de Monterrey, San Pedro Garza-García 66278, Mexico }
\date{}

\begin{document}
\maketitle

\begin{abstract}

Nanovectors (NVs), based on nanostructured matter such as nanoparticles (NPs), have proven to perform as excellent drug delivery systems. However, due to the great variety of potential NVs, including NPs materials and their functionalization, in addition to the plethora of molecules that could transport, this fields presents a great challenge in terms of resources to find NVs with the most optimal physicochemical properties such as particle size and drug loading, where most of efforts rely on trial and error experimentation. In this regard, Artificial intelligence (AI) and metaheuristic algorithms offer efficient of the state-of-the-art modelling and optimization, respectively. This review focuses, through a systematic search, on the use of artificial intelligence and metaheuristic algorithms for nanoparticle synthesis in drug delivery systems. The main findings are: neural networks are better at modelling NVs properties than linear regression algorithms and response surface methodology, there is a very limited number of studies comparing AI or metaheuristic algorithm, and there is no information regarding the appropriateness of calculations of the sample size. Based on these findings, multilayer perceptron artificial neural network and adaptive neuro fuzzy inference system were tested for their modelling performance with a NV dataset; finding the latter the better algorithm. For metaheuristic algorithms, benchmark functions were optimized with cuckoo search, firefly algorithm, genetic algorithm and symbiotic organism search; finding cuckoo search and symbiotic organism search with the best performance. Finally, methods to estimate appropriate sample size for AI algorithms are discussed.
\end{abstract}


\section{Introduction}

Nanotechnology is the manipulation of materials at the nanometric scale (typically 1-100 nanometers) with applications in multiple areas such as physics, chemistry, biology, design engineering, and medicine \cite{hyman2012,farokhzad2009,mitchell2020}. At the nanoscale several properties change, for example, the appearance of quantum effects,
the increase of the surface to volume ratio affecting the energy and elasticity of the nanomaterial \cite{N2019, Baer2018, Mozaffari2019}, the increased solubility as
particle size decreases \cite{Hochella2002}, changes in stable crystal structure \cite{Finnegan2007},
enhanced adsorption of molecules from the environment \cite{Jones2016}, particle growth and agglomeration \cite{Wu2011}. 

In medicine nanovectors (NV) have applications in different areas such as drug delivery, therapy, diagnostic, imaging and implants \cite{Wagner2006}.
For these applications the specific properties of the NPs, such as particle size and surface charge in conjunction with surface functionalization, make them suitable vehicles to transport desired molecules to a specific site while overcoming pharmacokinetic limitations associated
with conventional drug formulations \cite{Blanco2015}.

Although the industrial applications of nanotechnology are growing, they are not growing at the same rate as the advances in research and this could be in part due to reproducibility \cite{N2016}. The wide range of materials and methods that are used, and the absence of a database with sufficient amount of information can hinder the synthesis of the NVs because in order to obtain the desired properties, multiple experiments are needed to optimize the variables that are involved in the process, such as, concentration, volume fraction, temperature, contact time, sonication time, and voltage applied \cite{Tropsha2017}. However the lack of time, resources, and materials, can limit the number of experiments that are needed to reach a good set of properties, thus slowing down the pace of progress in nanotechnology either in terms of unoptimized products or outright hindering the search for new efficient NV designs.

To optimize the variables of NV synthesis, with or without molecule load, with a minimum number of experimental trials, the simplest way is to design experiments based on statistical methods so that a regression algorithm can model the properties using a first or second order polynomial equation. The most popular of such algorithms are: response surface methodology (RSM), partial least squares (PLS) and multiple linear regressions (MLR) \cite{Khuri2010,Wold2001}. However these methodologies rely on assumptions about the relationship between the input and the output variables, contrary to artificial intelligence (AI) algorithms \cite{fonner1970}. AI algorithms used for modelling are increasing in popularity because due to the lack of variable assumptions, these methods are less restrictive and are capable of recognizing and modelling non-polynomial relations between the variables \cite{Sun2003}. Examples of AI used for NV synthesis include multilayer perceptron artificial neural networks (MLP-ANN) \cite{Rouco2018,Li2015,Zaki2015}, adaptive neurofuzzy inference system (ANFIS) \cite{Ijadpanah-Saravi2017}, support vector machines (SVMs) \cite{Asfaram2016,Zhu2011}, genetic proggraming (GP) \cite{Azarhoosh2019,Tanzifi2020}, decision trees \cite{Dadrasi2019}, and random forests (RF) \cite{Azqhandi2017,Solaymani2017}.

Once the properties of a system are modelled, such as NV physicochemical properties, then the input variables of that system are optimized. 
In this regard, metaheuristic optimization is a versatile option because it does not require initial information about the process, and unlike most other optimization algorithms they are not gradient based, meaning that they do not require derivatives of the function that will be optimized. Most of metaheuristic algorithms are conceived as analogies of theories and behaviors studied by different areas such as biology, chemistry, physics and psychology \cite{Hussain2018}. Metaheuristic algorithms that have been used in processes involving nanomaterials are: genetic algorithms (GA) \cite{Tajmiri2020,Ghaedi2014Sulfide,Kumar2019}, particle swarm optimization (PSO) \cite{Ghaedi2014Methyl,Ijadpanah-Saravi2017,Zhu2011}, simulated annealing (SA) \cite{Hataminia2017}, differential evolution (DE) \cite{Feli2016,Lingamdinne2018}, ant colony optimization (ACO) \cite{Xu2019}, bees algorithm (BA) \cite{Ghaedi2015}, firefly algorithm (FA) \cite{kougianos2015}, cuckoo search (CS) \cite{khajeh2014}, bat algorithm \cite{ghaffarkhah2020}, gravitational search algorithm (GSA) \cite{Mohd2016} , symbiotic organism search (SOS) \cite{vinoth2017}, whale optimization algorithm (WOA) \cite{Dadvar2021} and grasshopper optimization algorithm (GOA) \cite{Dadvar2021}.
Deciding which algorithm to use is no easy task because the \textit{no free lunch} theorem for optimization states that algorithms which perform better for certain optimization problems will have a lower performance for other class of problems, and therefore there is no overall best algorithm \cite{Wolpert1997}.

In nanomedicine, several NV properties such as particle size (PS), shape, polidispersity index (PDI), zeta potential (ZP) and drug loading (DL) need to be optimized in drug delivery systems due to the biological barriers they encounter when applied in vivo or in vitro \cite{Blanco2015}.

Considering the wide range of AI modelling and metaheuristic optimization algorithms available for the study of nanomaterial properties, there are no reports describing and comparing them regarding NV synthesis in terms of PS, shape, PDI, ZP and DL as the optimized properties. The objective of this review is to provide an overview of the AI and metaheuristic optimization used to model NV properties focused on the aforementioned properties. Section 2 portrays the systematic methodology that was used for the literature search and algorithm evaluation. Sections 3 and 4 provide a brief description of AI and metaheuristic algorithms that have been used in NV synthesis. Section 5 compares AI algorithms by analyzing the conclusions of the articles that used two or more of these algorithms, as well as the results of a comparison of different metaheuristic algorithms by optimizing six benchmark functions. Section 6 gives a brief analysis of the articles that used AI and metaheuristic algorithms to model and optimize relevant NV properties for drug delivery systems. Finally, section 7 delves the challenge of sample size selection and its effects on the accuracy of results, as well as presents methods to assess such effects on AI algorithms performance.

\section{Methodology}\label{sec:Search}
A search in the Scopus and Web of Science databases was done to identify the existing literature of AI and metaheuristic algorithms that were used to model and optimize NP properties. The search was focused on variables used on drug delivery systems.

The set of keywords used in this systematic search were: 
\begin{itemize}
    \item TITLE-ABS-KEY ( (optimization OR modeling)  AND  algorithm  AND  ( nano  OR  nanoparticle )  AND  ( properties  OR  load  OR  synthesis ) )
\end{itemize}
A total of 864 entries were found on October 2020. After analyzing the abstracts it was concluded that out of the 864 entries only 133 modelled or optimized NP properties. The nanomaterial, composition, algorithms used, input variables and output variables (properties) from each of the 133 articles were collected. It is worth noting that the earliest entry is from the year 1999, however, out of the 133 articles from which the data was collected, the earliest article dates from 2007, reflecting that this area of research is relatively new.

\begin{figure}[H]
    \centering
    \includegraphics{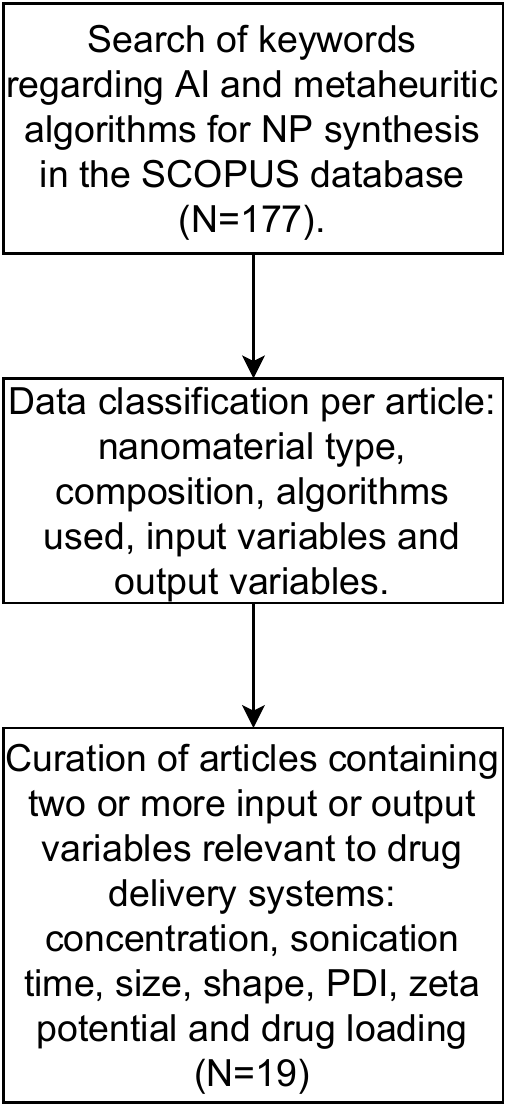}
    \caption{Steps of the systematic review process, N is the number of articles.}
    \label{fig:Met}
\end{figure}

Afterwards, the articles were analyzed looking for those that studied input or output variables that are of importance in drug delivery systems. Specifically, the input variables taken into consideration were concentration and sonication time, while the output variables were size, shape, PDI, ZP and DL. Of the 133 articles only 16 studied two or more of the input or output variables previously mentioned. A diagram of this systematic review process is presented in Figure \ref{fig:Met},
and Table \ref{tab:resumen2} presents the summary of the articles meeting the search criteria.

\begin{table*}[htp]
\centering
\begin{tabular}{@{}llllll@{}}
\toprule
\multicolumn{1}{c}{Material}                                         & \multicolumn{1}{c}{\begin{tabular}[c]{@{}c@{}}Modelling\\ Algorithm\end{tabular}} & \multicolumn{1}{c}{Input Variables}                                                                                                                                                                                                                                                                  & \multicolumn{1}{c}{Output Variables}                                  & \multicolumn{1}{c}{\begin{tabular}[c]{@{}c@{}}Sample\\  size\end{tabular}} & \multicolumn{1}{c}{Ref} \\ \midrule
ZrO2                                                                 & ANN-GA                                                                            & \begin{tabular}[c]{@{}l@{}}Concentration of an aqueous\\  solution of ZrOCl2·8H2O \\ (M{[}ZrOCl2·8H2O{]}), flow of \\ NH3 gas, and vacuum level \\ of the production system\end{tabular}                                                                                                                 & PS                                                                    & 19                                                                         &      \cite{Zhang2007}                   \\ \midrule
Agar                                                                 & ANN-GA                                                                            & \begin{tabular}[c]{@{}l@{}}CaCl2 (\%) \\ Homogenizer speed (rpm) \\ Agar (\%) HPCD (\%)\end{tabular}                                                                                                                                                                                                 & \begin{tabular}[c]{@{}l@{}}PS, PDI ZP, \\ DL, RE\end{tabular}         & 25                                                                         &   \cite{Zaki2015}                      \\ \midrule
Solid Lipds                                                          & ANN-GA                                                                            & \begin{tabular}[c]{@{}l@{}}Polymer content, eudragil RL, \\ drug content, Organic \\ phase type combined with\\ acetone\end{tabular}                                                                                                                                                            & RE                                                                    & 34                                                                         &      \cite{Varshosaz2012}                   \\ \midrule
Au                                                                   & ANN                                                                               &  \begin{tabular}[c]{@{}l@{}}LL\%, Tween 80\%, Lecithin\%, \\ Speed (rpm), Rifabutin\%\end{tabular}                                                                                                                                                                                                    & \begin{tabular}[c]{@{}l@{}}PS, PDI, ZP, \\ EF, DL\end{tabular}        & 29   &        \cite{Rouco2018}                  \\ \midrule
VRP                                                                  & SVM-GA                                                                            & \begin{tabular}[c]{@{}l@{}}Ratio of MAA to CS, \\ MAA/EGDMA and insulin\end{tabular}                                                                                                                                                                                                                 & \begin{tabular}[c]{@{}l@{}}PS, PDI, ZP, \\ DL and others\end{tabular} &                                       
& \cite{RoodbarShojaei2019}                        \\ \midrule
Carbon                                                               & ANN-GA                                                                            & \begin{tabular}[c]{@{}l@{}}Temperature, pH, HNO3 conc, and \\ sonication time\end{tabular}                                                                                                                                                                                                                  & PS                                                                    & 31                                                                         &    \cite{Holland1975}                     \\ \midrule
PLGA, PCL                                                            & ANN                                                                               & \begin{tabular}[c]{@{}l@{}}The density and hydrophilicity \\ of the shell polymers, feed rates\\ and concs of core and shell\\ phases, the contribution of \\ TCH in core material and \\ electrical field\end{tabular}                                                                              & \begin{tabular}[c]{@{}l@{}}Peppas \\ constants\end{tabular}           & 30                                                                         &                \cite{Maleki2014}         \\ \midrule 
PLN(VRP)                                                                  &     \begin{tabular}[c]{@{}l@{}}ANN-GA, \\ RSM\end{tabular}                            & \begin{tabular}[c]{@{}l@{}}Weight drug/lipid, and concentrations \\ of Tween 80 and Pluronic \\ F68\end{tabular}                                                                                                                                             &  DL, PS                                              & Test 14             & \cite{Li2015}                                                                              \\ \midrule
Magnetite                                                            & GA                                                                                & \begin{tabular}[c]{@{}l@{}}concs and compositions of \\ the copolypeptides, reaction \\ pH\end{tabular}                                                                                                                                                                                              & PS, shape                                                             &                                                                            &      \cite{Lenders2017}                   \\ \midrule
Ag                                                                   & \begin{tabular}[c]{@{}l@{}}ANN-PSO,\\ MPSO\end{tabular}                           & \begin{tabular}[c]{@{}l@{}}The Ratio of AgNO3/Opium, \\ feed rate (ml/min), pH, temp, \\ Agitation speed\end{tabular}                                                                                                                                                                                & PS                                                                    & 103                                                                        &          \cite{Shafaei2020}               \\ \midrule
Au                                                                   & \begin{tabular}[c]{@{}l@{}}GA \\ (SE)\end{tabular}                                & \begin{tabular}[c]{@{}l@{}}CTAB, HAuCl4 and AgNO3 \\ conc\end{tabular}                                                                                                                                                                                                                               & PS, Shape                                                             & 15                                                                         &      \cite{Salley2020}                   \\ \midrule
\begin{tabular}[c]{@{}l@{}}Carvedilol\\ nanosuspensions\end{tabular} & ANN-GA                                                                            & \begin{tabular}[c]{@{}l@{}}Drug load, stabilizer type, \\ stabilizer HLB, molecular\\ w. of stabilizer, numbers of\\ PEO and PPO, ratio of \\ hydrophilic to hydrophobic units,\\ w. of stabilizer, w. \\ drug /stabilizer, phase volume\\ aqueous/organic, and \\ aqueous phase volume\end{tabular} & PS, PDI, ZP                                                           & 36                                                                         &         \cite{Abdelbary2015}                \\
\hline
chitosan                     & ANN                                                                               & \begin{tabular}[c]{@{}l@{}}pH of polymer solution, conc\\ ratio of polymer/insulin and\\ polymer type\end{tabular}     & \begin{tabular}[c]{@{}l@{}}PS, ZP, PDS,\\  DL\end{tabular}            & 20                                                                         &      \cite{Shahsavari2014}                   \\ \hline
Au                           & \begin{tabular}[c]{@{}l@{}}GA\\ (SE)\end{tabular}                                 & \begin{tabular}[c]{@{}l@{}}Flux and ratio between Bn, \\ AgNO3 and NaOH\end{tabular}                                   & PDI                                                                   &                                                                            &    \cite{Fernandes2016}                     \\ \hline
Chitosan, plymer             & ANN                                                                               & \begin{tabular}[c]{@{}l@{}}TPP conc,chitosan concentrations,\\  chitosan: TPP mass ratios\end{tabular}                            & PS, ZP, DL                                                            & 45                                                                         &        \cite{Hashad2016}                 \\ \hline
Iron Oxide                   & \begin{tabular}[c]{@{}l@{}}SA-\\ LOOCV-\\ GRBF\end{tabular}                       & \begin{tabular}[c]{@{}l@{}}Temp, distilled water volume, ratio \\ of FA to iron moles, \\ adsorption time\end{tabular} & \begin{tabular}[c]{@{}l@{}}PS, FA \\ adsorption \\ value\end{tabular} & 16                                                                         &      \cite{Hataminia2017}                   \\
\bottomrule
\end{tabular}
\label{tab:resumen}
\end{table*}

\begin{table*}[h]
\centering
\begin{tabular}{@{}llllll@{}}
\toprule
\multicolumn{1}{c}{Material}                                         & \multicolumn{1}{c}{\begin{tabular}[c]{@{}c@{}}Modelling\\ Algorithm\end{tabular}} & \multicolumn{1}{c}{Input Variables}                                                                                                                                                                                                                                                                  & \multicolumn{1}{c}{Output Variables}                                  & \multicolumn{1}{c}{\begin{tabular}[c]{@{}c@{}}Sample\\  size\end{tabular}} & \multicolumn{1}{c}{Ref} \\ \midrule
Chitosan/streptokinase                  & ANN                    & \begin{tabular}[c]{@{}l@{}}pH, chitosan concentration \\ stir time\end{tabular} & \begin{tabular}[c]{@{}l@{}}PS, \\ Viability\end{tabular} & 30                                                                        &      \cite{baharifar2016cytotoxicity}                   \\

\midrule
PVA                   & \begin{tabular}[c]{@{}l@{}} ANN \end{tabular}                       & \begin{tabular}[c]{@{}l@{}} Polymer concentration,\\ Pressure, \\ PVA concentration\end{tabular} & \begin{tabular}[c]{@{}l@{}}PS, micropore \\ surface area\end{tabular}   & 36                                                                         &      \cite{rizkalla2005}   \\

\bottomrule
\end{tabular}
\caption{Summary of the characteristics of articles studying drug delivering systems satisfying the conditions stated in section \ref{sec:Search}.}
\label{tab:resumen2}
\end{table*}

\section{AI  algorithms}
AI is a branch of computer science that focus on giving to machines, devices or programs the ability to show attributes of intelligence such as cognitive capacity, memory, learning, and decision making \cite{chen2019, Feldman2001}. AI has been applied in areas such as pharmaceutics \cite{gawehn2016}, computer vision \cite{Lecun2015}, sustainable energies \cite{Zhao2012}, genetics \cite{Libbrecht2015}, medical imaging \cite{Shin2016}, material science \cite{Gomez-Bombarelli2016}, among others.

Within AI, Machine learning is a subset that uses inductive reasoning, \textit{i. e.} arrives at a general conclusion learning from particular examples \cite{Michie1968}. Machine learning algorithms include neural networks, SVM, decision trees, RF and GP. In what follows, machine learning algorithms used for NP optimization are described.

\begin{quote}
    \textit{Multilayer perceptron artificial neural network (MLP-ANN)}
\end{quote}

MLP-ANN is the most used AI algorithm for modelling, predicting, and forecasting. ANN is a mathematical model inspired by the nervous system, in which biological neurons are connected to other neurons by the dendrites and axons, between these elements lies the synapse, thorugh which passes a flow of electrical signals that contain information \cite{McCulloch1943}. 

ANNs structure consists of three types of layer: the input, the hidden and the output layers. In the input layer the nodes, representing the neurons, are the input variables and their values. The input of the hidden layer is represented by the sum of the inputs associated to a node, multiplied by their weights, representing the synapse as the links between the neurons. Equation \ref{eq:1} represents this mathematically:
\begin{equation}\label{eq:1}
    H_{in} = \sum_{i}^N w_i x_i + r\,,
\end{equation}
where $H_{in}$ is the input of the neuron of the hidden layer; $N$ is the number of inputs associated with that neuron; $w_i$ are the associated weights to the synapse; $x_i$ are the input variables; and $r$ is an adjustable parameter.
The output of the hidden layer is given by the activation function
\begin{equation}
    H_{out} = A(H_{in}) \,,
\end{equation}
where $A(x)$ is the activation function. The most common activation functions for the hidden layers are \cite{Glorot2011}
\begin{itemize}
    \item Sigmoid
    \begin{equation}\label{eq:Act_sigmoid}
        A(x) = \frac{1}{1+e^{-x}}
    \end{equation}
    
    \item Hyperbolic tangent
    \begin{equation}\label{eq:Act_tangent}
        A(x) = \frac{e^{x} - e^{-x}}{e^{x}+e^{-x}}
    \end{equation}
    
    \item Threshold
    \begin{equation}
        A(x) = \begin{cases} 
      0 & x\leq 0 \\
      1 & x>0 
   \end{cases}
    \end{equation}
    
    \item Rectified Linear unit (RELU)
    \begin{equation}
        A(x) = \max\{0,x\}
    \end{equation}
\end{itemize}
with RELU being the most used in recent years, due to its computational efficiency, given that it does not need to compute the exponential function \cite{Krizhevsky2017}. 

Most of MLP-ANN are constructed with only one hidden layer. However, there are special applications, such as speech recognition \cite{Hinton2012} and image recognition \cite{Krizhevsky2017}, whose use more than 3 hidden layers. These are called deep neural networks.

In the output layer, the input is the sum of all the outputs of the hidden layer, each multiplied by a weight, and finally adding some bias, as described by the following equation:
\begin{equation}
    O_{in} = \sum_{i}^N w_i H_{out,i} + r\,,
\end{equation}
where $O_{in}$ represents the input of the output layer; N is the number of neurons in the last hidden layer; $w_i$ are the associated weights of the synapses; and $r$ is an adjustable parameter.
The most common activation function for the output layer is the linear function $y=x$ so the output is given by
\begin{equation}
    O_{out} = O_{in}
\end{equation}

After the outputs are calculated, they are compared with the dependent variables that are going to be modelled. For this comparison an error function is defined
\begin{equation}
    E_{tot} = \sum \frac{1}{2} (target - O_{out})^2 \,,
\end{equation}
where $target$ is the actual dependent variable, and $O_{out}$ is the modelled dependent variable.

After the errors are calculated now the weights of the synapses are updated to minimize the error through an algorithm called gradient descent which consists on the following equation: 
\begin{equation}
    w_i(t+1) = w_i(t) - \eta\frac{\partial E_{tot}}{\partial w_i}
\end{equation}
where $\eta$ is called the learning rate, a paramater defined by the user which often take values in the range [0.1,0.9]. The error does not explicitly depends on the weights, so the algorithm needs to go 'backwards' and calculate a series of partial derivatives with respect to previous function to obtain $\partial E / \partial w_i$, this process is called backpropagation. 

\begin{figure}
    \centering
    \includegraphics[scale=0.27]{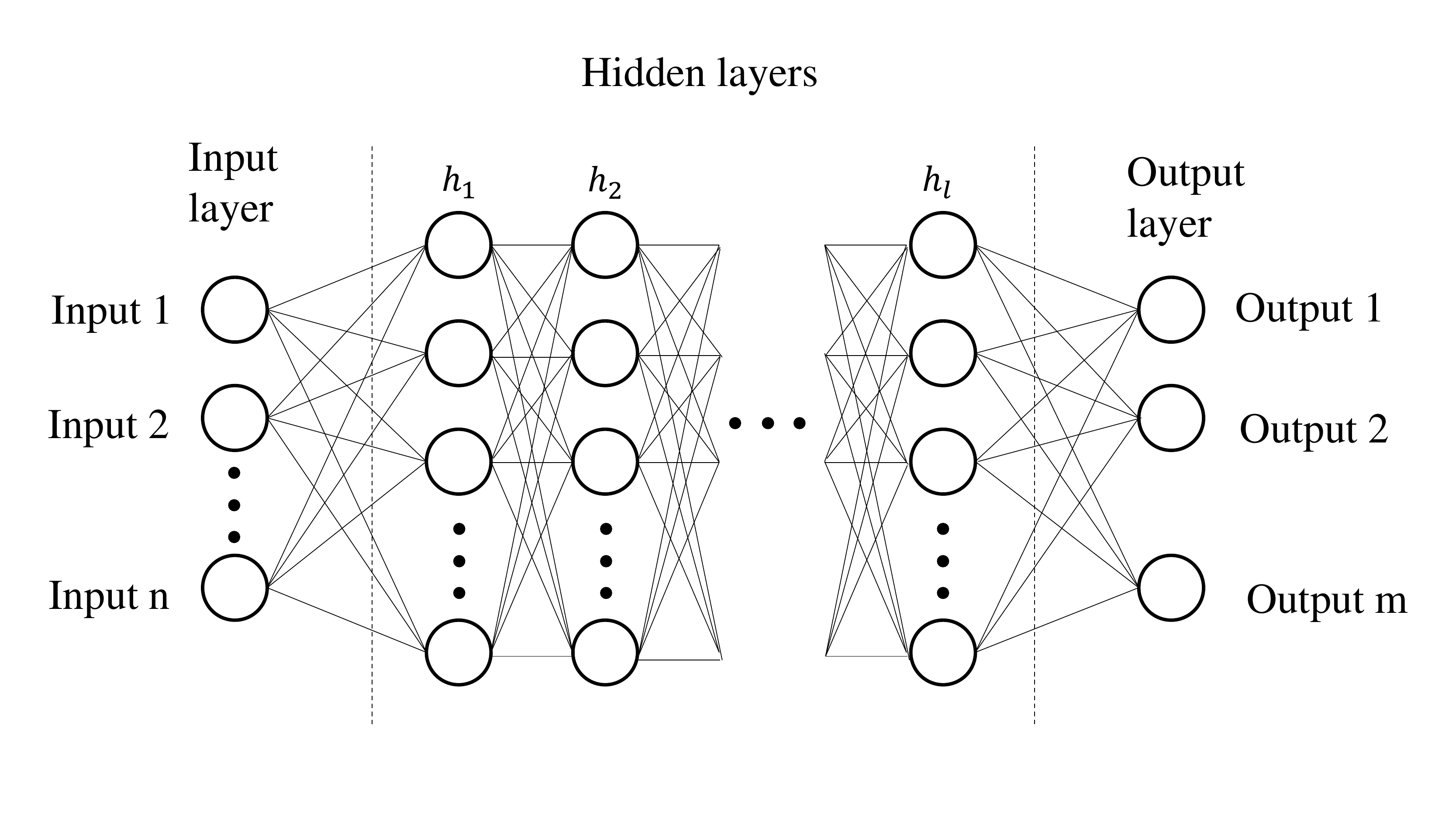}
    \caption{General form of an MLP-ANN model.}
    \label{fig:ANN_MLP}
\end{figure}

The training by gradient descent is easy to implement but has a very low speed of convergence. Other training algorithms which can be faster but more complex to implement are Levenberg-Marquadt, Bayesian regularization, conjugate gradient, quasi-Newton, secant, and scaled conjugate gradient \cite{Hagan1994, Burden2008, Moller1993}. The general form for an MLP-ANN is shown in figure \ref{fig:ANN_MLP}.

\begin{quote}
    \textit{Radial basis function neural network (RBF-NN)}
\end{quote}
RBF-NN are composed of the same type of layers as MLP-ANN, and the structure is the same, the main differences lies in the activation functions in the hidden layer (being the Gaussian the most comon one in RBF-NN), and that each neuron in the hidden layer represents a data cluster centered at a specific point within a given radius \cite{Hemmati-Sarapardeh2018,Yu2008,BROOMHEAD1988}.

\begin{quote}
    \textit{Support Vector Machines (SVM)}
\end{quote}

A SVM is a computer algorithm that learns by observing examples and then assign labels to objects \cite{Boser1992}. SVM was initially designed to solve pattern recognition problems trying to find a rule by analyzing a sample of input variables called support vectors. SVM maps the input variables $x \in \mathbb{R}^n$, where $n$ is the number of input variables, to a space with a higher dimensionality $m>n$, this is done because data can be better classified by a function, called kernel, defined in the space $\mathbb{R}^m$. The kernel function defines a hyper plane that  maximizes the distance from any given data point between different classes and therefore the plane is capable of classify the data points \cite{Sain1996}.

\begin{quote}
    \textit{Decision trees}
\end{quote}

Decision trees can predict adequately the value of an output variable by learning decision rules inferred from the input variables. Decision trees have the structure of a tree with the most general diversion path beginning in the roots, then the trunk, branches and ending in the leafs. A node represents a test case and the ramifications are the possible solutions to that case. Methods like ID3, ASISTANT and C4.5 have been developed to train a decision tree, with ID3 being the easier to implement \cite{Quinlan1987, Mitchell1999}.


\begin{quote}
    \textit{Random forests (RF)}
\end{quote}
RF consist on several decision trees that are combined by calculating a weighted mean. Each tree is constructed with different training sets and unlike in a single decision tree model, the trees are allowed to grow to their maximum. The parameters are defined by the user are the number of trees in the forest, the number of candidate random variables in each ramification, the percentage of the data that is going to be used for training, generally a value between 60 and 80\%, the minimum number of samples in the terminal neurons and the maximum number of terminal neurons \cite{Breiman2001}.

\begin{quote}
    \textit{Adaptive Neuro Fuzzy Inference System (ANFIS)}
\end{quote}
ANFIS builds a series of fuzzy if-then rules using the structure of a neural network \cite{jang1993}. The architecture of ANFIS consists on five layers. Layers 1 and 4 have variable nodes (neurons), meaning that there are parameters that need to be fitted. 

The output of the first layer assigns the variables to a linguistic label by calculating the degree of membership using the following equation
\begin{equation}
    O_i^1 = \mu_{A_{i}} (x) \,,
\end{equation}
where $x$ is the input variable, $A_i$ is the linguistic label and $\mu_{A_{i}}$ is the membership function. Bell shaped membership functions are often used, like Gaussian (eq. \ref{Gau}) or Lorentzian (eq. \ref{Lor}) functions:
\begin{equation}
    \label{Gau}
    \mu_{A_{i}} (x) = \frac{1}{1 + [\frac{x-c_i}{a_i}]^{2b_i}}
\end{equation}

\begin{equation}
    \label{Lor}
    \mu_{A_{i}} (x) = \exp{-(\frac{x-c_i}{a_i})^2}
\end{equation}
where $a_i$, $b_i$ and $c_i$ are called premise parameters. In the second layer, the output of the membership function is multiplied over all the possible combinations of linguistic labels. This is represented by the $\omega$ letter and is called the firing strength.
\begin{equation}
    \omega_i = \prod_{j=1}^N \mu_{A_i^{(j)}}(x^{(j)})
\end{equation}
where $N$ is the number of input variables. The third layer is the normalization of the firing strengths, i.e.
\begin{equation}
    \bar{\omega_i} = \frac{\omega_i}{\sum_{i=1}^N \omega_i} \,.
\end{equation}
The fourth layer is an adaptive layer where the output is given by
\begin{equation}
    O_i^4 = \bar{\omega_i} f_i\,,
\end{equation}
where $f_i$ is a linear function of the input variables
\begin{equation}
    f_i = \sum_{j=1}^M  p_{i,j} x_{i,j} + r_i \,,
\end{equation}
where M is the number of input variables; $p_j$ and $r$ are parameters called consequence parameters. The fifth and last layer consists of the sum of all the outputs in the fourth layer
\begin{equation}
    O^5 = \sum_i \bar{\omega_i} f_i\,.
\end{equation}

\begin{figure}
    \centering
    \includegraphics[scale=0.27]{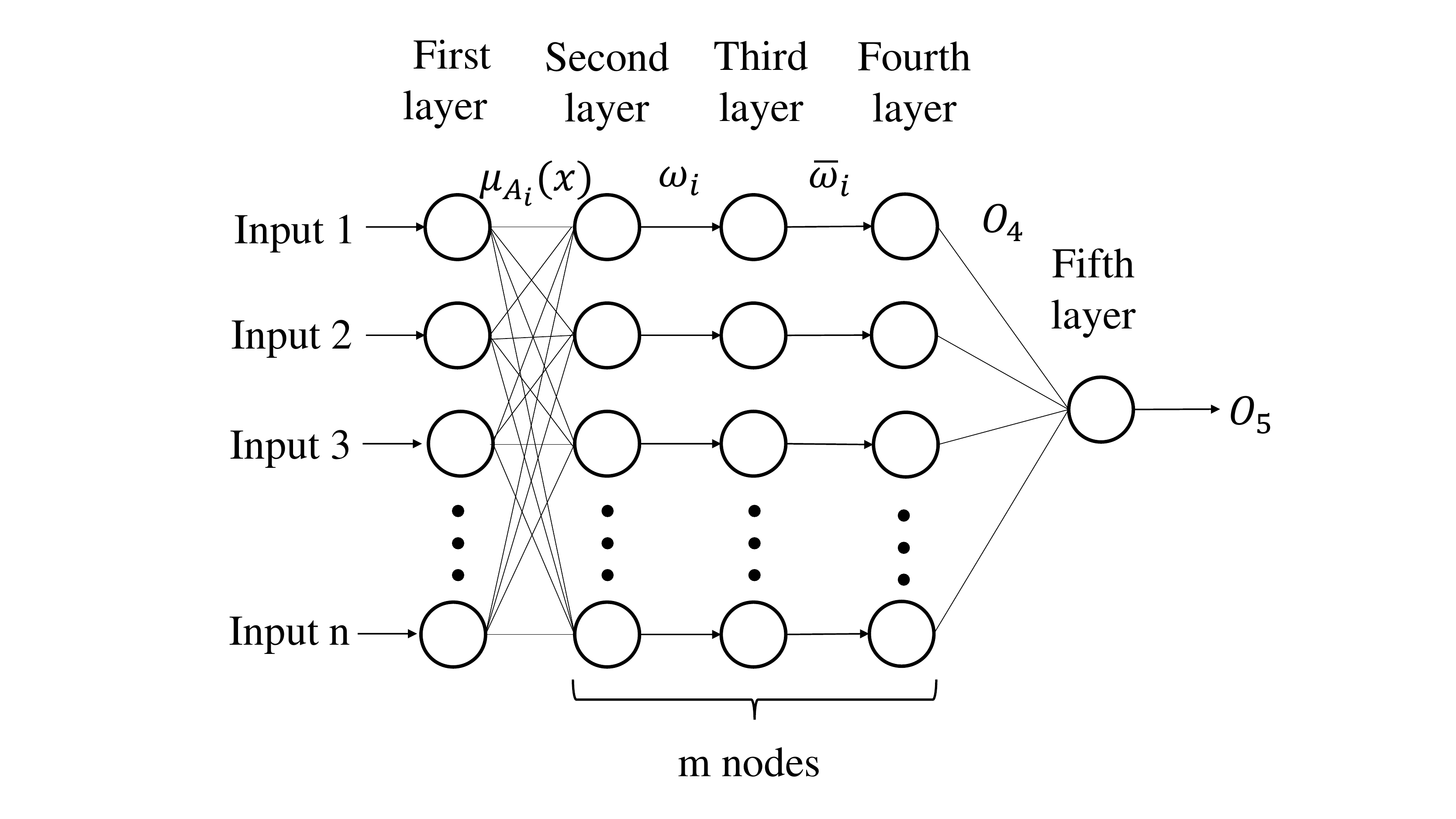}
    \caption{Schematic of an ANFIS structure}
    \label{fig:ANFIS}
\end{figure}

An overview of the internal ANFIS structure is shown in figure \ref{fig:ANFIS}.

\begin{quote}
    \textit{Genetic Programming (GP)}
\end{quote}

GP is a regression algorithm that uses the Darwin evolution theory for the selection of functions to model a phenomena. GA was the basis algorithm for the development of GP, however the former optimizes the input numerical variables in an already defined or modelled space, while the latter generates a function
that models the output variables. GP uses operations such as crossover, mutation, reproduction, gene duplication, and gene deletion to find the most suitable function that adjust the given data \cite{Koza1992, poli2014}.

\section{Metaheuristic optimization}
The word metaheuristic comes from the greek \textit{meta} which means to go further and \textit{heuristic} which means find or create. Metaheuristics is a heuristic method to solve a computational problem, particularly used for optimization.

\begin{figure}[H]
    \centering
    \includegraphics[scale=0.4]{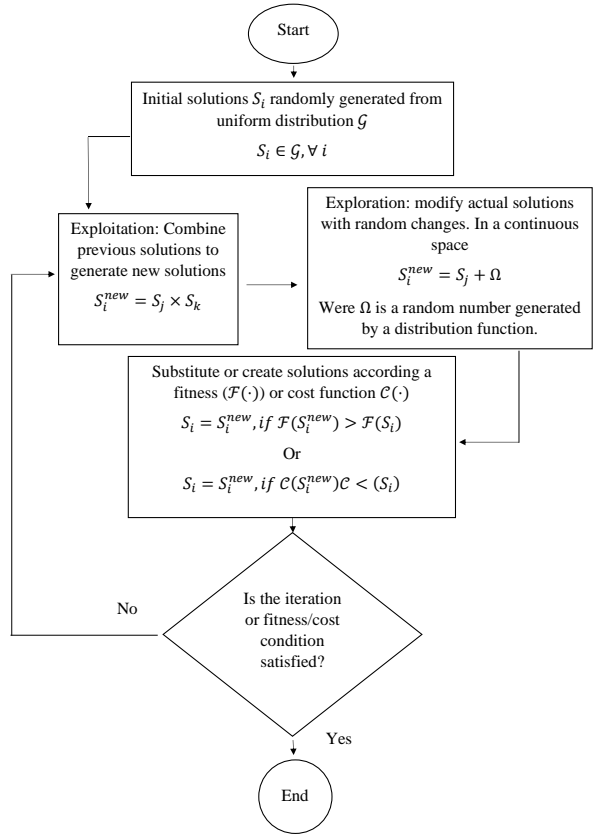}
    \caption{General overview for a metaheuristic optimization algorithm.}
    \label{fig:metaheuristic_diagram}
\end{figure}

There are different criteria to classify the metaheuristic algorithms, however one of the most used is the number of solutions that are proposed. Individual search algorithms, like SA and taboo search, use only one solution per iteration while population search algorithms, like GA and PSO use multiple solutions per iteration \cite{birattari2001}. Population search algorithms are more popular in the research field due to their robustness, and are characterized by an exploitation phase or local search and an exploration phase or global search. In contrast, single solution algorithms usually focus on local search, thus population search algorithms tend to avoid local optima compared to individual search algorithms \cite{Mirjalili2016, Boussaid2013}. 
Most of metaheuristic algorithms use a function to evaluate the performance of the solutions that are generated. In the literature these functions have different denominations: cost function, fitness function or objective function. The cost is associated with a minimization problem, the fitness is associated with a maximization problem, and the objective function can be used for both. In the following algorithms the notation OF (objective function) and OFV (objective function value) will be used for algorithms that accept both types of optimization, also the term 'fittest' will be used to refer to the best solutions found by the algorithm if both types of optimization are accepted regardless of whether a cost or fitness function was used. Another common feature of metaheuristic algorithms is that many of them use parameters that need to be defined by the user. Values or ranges for these parameters can be recommended, however, there is no specific value that yields an optimum performance, turning parameter tuning into a complicated process. Finally, in all the algorithms the initial population, or individual, is generated 
at random within the search space defined by the user, a general structure of these algorithms is shown in figure \ref{fig:metaheuristic_diagram}. In what follows, metaheuristic algorithms used for NP optimization are described.

\begin{quote}
    \textit{Genetic Algorithm (GA)}
\end{quote}
GA is one of the most used metaheuristic algorithms. The most basic one consists on three steps: natural selection, crossover and mutation \cite{Holland1975}. Natural selection consists on choosing the fittest candidates for survival. There are two ways to carry out the natural selection: by roulette or by tournament, in the former all individuals are ordered by their OFV and in the latter pairs of individuals are chosen randomly and the fittest individual survives. Crossover consists in combining pairs of individuals from the last generation to create new individuals, an analogy between parents and children. In a GA with continuous search space the crossover is given by the following equations
\begin{equation}
    x_{i}(t+1)=x_i(t) - rand(0,1)\times[x_i(t) - x_j(t)]\,,
\end{equation}
\begin{equation}
    x_{j}(t+1)=x_j(t) + rand(0,1)\times[x_i(t) - x_j(t)]\,,
\end{equation}
where $x_i$ and $x_j$ are individuals chosen randomly and $rand(0,1)$  is a vector of random numbers between 0 and 1. In GA not all individuals execute the crossover but rather there is a probability called crossover rate. This rate is a parameter defined by the user, in general it is set in a range [0.7,0.9]. Mutation consists on changing completely the individual in some number of chromosomes (dimensions) chosen at random, but leaving equal the rest. The chromosomes that are changed take a random value in the search space so the solution does not get trapped in a local minimum. The mutation operation is given by the following equation 
\begin{equation}
    x_{mut} = rand(0,1)\times(UB-LB) +LB\,,
\end{equation}
where $rand(0,1)$ is a vector of random numbers between 0 and 1; UB and LB are the upper bound limit and the lower bound limit of the search space respectively. Similar to the crossover rate, the mutation rate is the probability that the operation takes place and is defined by the user. After the mutation step, the process is repeated from the beginning until a number of generations (iterations) is reached or until an individual reaches a certain OFV. 

\begin{quote}
    \textit{Cuckoo Search (CS)}
\end{quote}

CS is based on the breeding behaviour of the cuckoo birds. The cuckoo birds are brood parasites that lay their eggs in another nest of birds, the birds are born and raised by other bird species \cite{Yang2009cs, Kruger2002}. For the algorithm it is implied that a cuckoo lays only one egg in the nest of the other bird, and the host bird has a probability $p_a$ of discovering the cuckoo egg and abandon it.

The initial randomly generated solutions $x_i(0)$ are called eggs. The cuckoo lays a new egg in the nest via Levy flight. This new solution is given by
\begin{equation}
    x_i(t+1) = x_i(t) + \alpha \times Levy(\lambda)
\end{equation}
where $\alpha$ is a parameter related to the stepisize, in most cases it can be used as 0.01 \cite{Yang2009cs}; and $Levy(\lambda)$ is a number drawn from the Levy distribution which is given by
\begin{equation}
    Levy = t^{-\lambda}\,,
\end{equation}
where $\lambda$ takes a value between 1 and 3. The cuckoo egg replaces the host egg $x_i(t)$ only if the OFV is better for the cuckoo egg. After the eggs are replaced, the egg with the worst OFV is abandoned with a probability $p_\alpha$, the new egg changes in some dimensions by a crossover operation and the remainder of the dimensions are left as they were. The new dimensions of the solution are given by
\begin{equation}
    x_i(t+1) = x_i(t) + rand(0,1)\times(x_{r1}(t) - x_{r2}(t))
\end{equation}
where $x_{r1}$ and $x_{r2}$ are eggs chosen randomly. The algorithm begins a new iteration by laying new cuckoo eggs until a termination criteria is reached.

\begin{quote}
    \textit{Symbiotic Organism Search (SOS)}
\end{quote}

SOS simulates the symbiotic interaction between organisms. The set of all the initial proposed solutions is called ecosystem and each individual solution is called organism. SOS consists on three steps: mutualism, commensalism and parasitism. In SOS the iterations $i$ go from 1 to N, where N is the size of the ecosystem and then the three steps take place.

Mutualism is a relationship between two organisms in which both of them are benefited, an example is the relationship between the bees and the flowers. The benefit is represented in the following equations
\begin{equation}
    x_i^{new} = x_i + rand(0,1)\times[x_{best} - MV\times BF_1]\,,
\end{equation}
\begin{equation}
    x_j^{new} = x_j + rand(0,1)\times[x_{best} - MV\times BF_2]\,,
\end{equation}
\begin{equation}
    MV=(x_i + x_j)/2\,,
\end{equation}
where $x_i$ is the organism of the current iteration; $x_j$ is an organism chosen at random from the total population; $BF_1$ and $BF_2$ are the benefit factor which can take a value of 1 or 2, this factors represent how much the organism is benefited; MV is the mutual vector which represent the relationship between the organisms; and $x_{best}$ is the fittest organism in the ecosystem.
Finally, the organisms are updated only if the OFV of the new organisms is better than that the current organisms. Comensalism is a relationship between two organisms in which only one is benefited while the other organism is not affected. An example of comensalism is the relationship between the shark and the remora fish. In SOS comensalism is represented by the following equation
\begin{equation}
    x_i^{new} = x_i + rand(-1,1)\times [x_{best} - x_{j}]\,.
\end{equation}
As in mutualism, the organism is updated only if the OFV of the new organism is better than the OFV of the current organism. Parasitism is a relationship between two organisms in which one organism is benefited while the other is affected in a negative way. An example of parasitism is the relationship between the anopheles mosquito and the humans. In SOS a parasite vector is created from one of the organisms changing some dimensions of the organism at random. The equation of the parasite vector is
\begin{equation}
    x_{par} = rand(0,1)*(UB-LB) + LB\,,
\end{equation}
the $x_{par}$ substitutes a randomly chosen organism from the ecosystem only if the OFV of the parasite vector is better.

After the parasitism phase is over then the three processes are applied to the following organism $i=i+1$ and when all the ecosystem goes through the three steps then the OFV of each individual from the new ecosystem is measured and the process begins again until an individual meets a criteria of solution or until a number of ecosystems are reached.

SOS has an advantage over most of metaheuristic algorithms in that no parameters have to be defined by the user. Due to this SOS is gaining popularity on optimization problems of engineering design, communication networks, machine learning, among others \cite{Ezugwu2019}.

\begin{quote}
    \textit{Firefly algorithm (FA)}
\end{quote}

FA is based on the behaviour of the fireflies, female fireflies can emit light to attract mates for reproduction. The higher the intensity of the glow, the higher the probability of finding a mate. In FA each solution represents a firefly and each firefly has a glow intensity which depends on the OFV, and the the interactions between the fireflies is determined by their intensity and the distance between them. The movement of fireflies with less glow intensity $i$ to the ones that have more glow intensity $j$ is modelled by the following equation:
\begin{equation}
    \bf{x_i} = \bf{x_i} + \beta\exp{-\gamma r_{ij}^2}(\bf{x_j} - \bf{x_i}) + {\alpha}\epsilon_i \,,
\end{equation}
where $r_{ij}$ is the distance between the fireflies $i$ and $j$; $\epsilon_i$ represents a random vector within the search space; and $\alpha, \beta, \gamma$ are parameters defined by the used. Suggested ranges for the parameters are $\alpha \in [0,1]$, $\beta \in [0.8, 1.2]$ y $\gamma \in [0,1]$ \cite{Yang2009}.

\begin{quote}
    \textit{Particle Swarm Optimization (PSO)}
\end{quote}

PSO, like GA, is one of the most popular metaheuristic algorithms for optimization. PSO is based on a simplified social model which uses the metaphor of the flocking behavior of birds to solve optimization problems \cite{Eberhart1995}.

The initial group of proposed solutions is called swarm and each individual solution is called particle. Then the particles start to move with a velocity and the positions are updated by the equation of motion $x(t+1)=x(t) + v(t)$, where $x$ is the position, $v$ is the velocity, and $t$ is the number of iteration. 
The process is repeated until an OFV is met or until a certain number of iterations are reached.

The advantage of PSO over GA is that when particles are near of the optimum solution they do not change abruptly, unlike in GA when the crossover takes part. Another advantage is that particles have memory so if a particle fly past some optima, the particle tends to return towards the solution. However, the standard PSO tends to converge on local optima in some type of problems and to solve this several modified versions of PSO have been proposed \cite{Pant2009}.

\begin{quote}
    \textit{Simulated Annealing (SA)}
\end{quote}

SA was one of the first metaheuristic algorithms that were developed. SA is a single solution algorithm that is based on the metropolis algorithm from statistical mechanics which search the most probable configuration of spins in a ferromagnetic material taking into account the Boltzman distribution and the partition function $Z = \sum \exp{E/k_B T}$ \cite{Metropolis1953, Kirkpatrick1983}.


\begin{quote}
    \textit{Gravitational Search Algorithm (GSA)}
\end{quote}

GSA is inspired by the Newton's law of gravity \cite{Rashedi2009}, this law states that two or more objects experiment an attractive force directly proportional to the masses of the objects and inversely proportional to the distance between them.

The individual solutions that are generated are called objects, a cost function is defined, this function determines the mass of each object. Next, the objects start to move toward each other, with less massive objects moving faster towards the more massive ones. After the new positions are updated, the mass of each object is calculated and the process begins again. The algorithm is stopped when all the objects are in the same position or when a certain number of iterations are reached.

\section{Comparison of algorithms}

From the 133 articles that were analyzed, 29 of them compared the performance of 2 or more modelling algorithms. Table \ref{tab:Comparacion} shows the comparison of the algorithms and the number of articles in which the comparison was made. 
To compare quantitatively the errors and performance of the algorithms, the authors used several methods: coefficient of determination ($R^2$), mean absolute error (MAE), mean percentage error, mean squared error (MSE), sum of the residuals,  variance accounted for, among others. However, the most common ones were $R^2$ and MSE, in fact only 2 of the 28 articles did not use $R^2$ to measure the performance.

In all articles that used AI or RSM the algorithms performed well. In 2 articles MLR the performance was deficient due to the restriction of adjusting the parameters to a linear equation \cite{Ghaedi2014Methyl, Azarhoosh2019}. Although both AI and fitting equation methods obtained good results in most articles, the former ones showed better performance than the later ones. The most common comparison was between MLP-ANN and RSM with 16 articles comparing them, and MLP-ANN outperformed RSM everytime. 
From this qualitative assessment, it is clear that AI algorithms fare better for modelling NP properties.

\begin{table}[H]
\centering
\begin{tabular}{c|c}
Comparison                   & No. of articles \\ \hline
MLP-ANN$>$RSM                & 13                 \\
MLP-ANN$>$MLR                & 6                  \\
MLP-ANN$>$RSM$>$GP           & 1                  \\
MLP-ANN$>$PLS$>$RSM          & 1                  \\
MLP-ANN$>$Decision trees$>$RSM & 1                \\
MLP-ANN$>$RF$>$RSM           & 1                  \\
MLP-ANN$>$PR                 & 1                  \\
ANFIS$>$MLP-ANN              & 1                  \\
RF$>$RSM                     & 1                  \\
RF$>$RBF-NN$>$RSM            & 1                  \\
SVM$>$MLP-ANN$>$RSM          & 1                  \\
SVM$>$RSM                    & 1                  \\
\end{tabular}
\caption{Comparison of modelling algorithms in the 28 articles. $x_1>x_2$ means that algorithm $x_1$ showed a better predictive capacity of the properties than the algorithm $x_2$. PR : Polynomial regression. \cite{Tanzifi2020, Tajmiri2020, Farhadi2020, Moghri2016, Ghaedi2014, Dadrasi2019, GhanavatiNasab2018, Solaymani2017, Azqhandi2017, Mehrabi2017, Ijadpanah-Saravi2017, Asfaram2016, Li2015, Ghaedi2015, Mousavi2013, Varshosaz2012, Zhu2011, Azarhoosh2019, Kumar2019, Daraei2014, Nasab2019, Hou2020, Saha2018, Mohan2015, Ghaedi2014Methyl, Ghaedi2014Sulfide, Karri2018, Zaki2015, rizkalla2005}. }
\label{tab:Comparacion}
\end{table}



For optimization algorithms comparisons were scarce,
only 4 articles compared metaheuristic algorithms. GA outperformed PSO in optimizing the concentration, sonication time, pH and amount of adsorbent for the removal of ethyl violet in wastewater \cite{Hou2020}, and outperformed GSA and PSO in optimizing parameter combination among selected datasets for ZnO synthesis \cite{Norlina2015}. In other words, PSO outperformed GA in optimizing the solid concentration and temperature for the removal of crystal violet using bimetallic Fe/Ni NPs \cite{Ruan2018}, and in training an ANFIS network to predict the thermophysical properties of Al2O3-MWCNT/oil hybrid nanofluid \cite{Alarifi2019}.

Because of the lack of information obtained about which optimization algorithm could be better, four algorithms were chosen to be compared: GA, SOS, FA and CS. GA was chosen because it is the most used metaheuristic algorithm for optimization; SOS because it has no parameters to be tuned by the user; FA due to simple implementation; and CS because of its unique Levy distribution, which helps the algorithm to avoid local optima. The comparison consisted on implementing the 4 algorithms to find the minimum of 6 benchmark functions, measuring the accuracy, precision and time of convergence of each algorithm.

The benchmark functions were
\begin{itemize}
    \item [1)] Schaffer N1
    \begin{equation}
        f(x, y)=0.5 + \frac{\sin^2{(x^2+y^2)^2}-0.5}{(1+0.001(x^2+y^2))^2}
    \end{equation}
    
    \item [2)] Holder-Table
    \begin{equation}
        f(x,y)=-\left|\sin x\cos y\exp(\left|1-\frac{\sqrt{x^2+y^2}}{\pi}\right|)\right|
    \end{equation}
    
    \item [3)] Cross-in-Tray
    \begin{equation}
    \begin{array}{ll}
        f (x,y)=-0.0001(|\sin(x)\sin(y) \\ \exp(|100-\frac{\sqrt{x^2+y^2}}{\pi}|)|+1)^{0.1}
        \end{array}
    \end{equation}
    
    \item [4)] Happy cat
    \begin{equation}
        f(\textbf{x})=\left[\left(|\textbf{x}|^2 - n\right)^2\right]^\alpha + \frac{1}{n}\left(\frac{1}{2}|\textbf{x}|^2+\sum_{i=1}^{n}x_i\right)+\frac{1}{2}
    \end{equation}
    
    \item [5)] Schaffer N4
    \begin{equation}
        f(x, y)=0.5 + \frac{\cos^2(\sin{\abs{x^2+y^2}})-0.5}{(1+0.001(x^2+y^2))^2}
    \end{equation}
    
    \item [6)] N-D Sphere
    \begin{equation}
        f(\textbf{x})= \sum_{i=1}^n x_i^2
    \end{equation}
\end{itemize}

In table \ref{DF} a more detailed description of the functions can be found.

\begin{table}
\centering
\begin{tabular}{c|ccccc}
Function       & Domain            & Dimensions & Type    & $f(x_{min})$ & $x_{min}$                                                        \\ \hline
Schaffer N1   & {[}-50,50{]}     & 2           & D,NS,U  & 0            & (0,0)                                                                  \\
Holder-Table  & {[}-10,10{]}     & 2           & ND,NS,M & -19.2085     & ($\abs{8.05502}$,$\abs{9.66459}$)                     \\
Cross-in-Tray & {[}-10,10{]}     & 2           & ND,NS,M & -2.2795      & ($\abs{1.349406}$,$\abs{1.349406}$) \\
Happy cat     & {[}-2,2{]}       & 2           & D,NS,M  & 0            & (-1,-1)                                                                \\
Schaffer N4   & {[}-50,50{]}     & 2           & D,NS,U  & 0.292579     & (0,1.253115)                                                           \\
6D sphere     & {[}-5.12,5.12{]} & 6           & D,S,U   & 0            & (0,0,0,0,0,0)                                                         
\end{tabular}
\caption{Benchmark functions. The second column describes the boundaries of the search space that was used for each function. The third column is the number of dimensions. The fourth column gives a description of the function, D: differentiable; ND: non-differentiable; NS: non-separable; U: unimodal; M: multimodal. $f(x_{min})$ is the minimum of the function and $x_{min}$ is where the minimum can be found in the domain of the function, the $\abs{x}$ symbol means that the value of $x$ can be positive or negative.}
\label{DF}
\end{table}

A population of 50 was used for each algorithm, for GA the crossover and mutation rates were 0.9 and 0.01 respectively, for FA the value of the parameters were $\alpha = 0.2$, $\beta = 1$ and $\gamma = 1$, and for CS $Pa = 0.25$. These values were used because either the author of the algorithm proposed those values or because they are the most common ones \cite{Yang2009, Yang2009cs}.

Each algorithm was tested 6 times for each function, obtaining the mean and standard deviation for each test as shown in table \ref{tab:exacto}. It can be noted that SOS and CS were the only algorithms that could find the minimum of the Schaffer N2 function and that all algorithms reached the minimum for the non-differentiable functions Holder-Table and Cross-in-Tray. However none of them found the minimum of the Happy-Cat function, where CS showed a slightly better performance than the other algorithms, but all of them obtained similar results in terms of accuracy and precision. 

The results of visual convergence time are shown in table \ref{tab:tiempo}. For this table the algorithms were run 10 times between 0.06 and 0.08 seconds and the OF was plotted against the running time. Then a point in time was taken for each algorithm where the algorithm converged visually (i.e. the OF did not seem to change after that point in time). GA was by far the fastest algorithm to have converged followed by SOS and CS, while FA did not converged in the given set of times for any function.

\begin{table*}[ht]
\centering
\begin{tabular}{@{}ccccccc@{}}
\toprule
function                                                                 &                 & Solution          & GA                     & SOS                 & FA         & CS                  \\ \midrule
\multirow{4}{*}{\begin{tabular}[c]{@{}c@{}}Schaffer \\ N2\end{tabular}}  & $[x]$           & {[}0{]}           & -0.00002               & \textbf{0}          & 0.24988    & \textbf{0}          \\
                                                                         & $\sigma_x$      & NA                & 0.00004                & \textbf{0}          & 1.06468    & \textbf{0}          \\
                                                                         & $f(x)$          & 0                 & 0                      & \textbf{0}          & 0.00208    & \textbf{0}          \\
                                                                         & $\sigma_{f(x)}$ & NA                & 0                      & \textbf{0}          & 0.00161    & \textbf{0}          \\ \midrule
\multirow{6}{*}{\begin{tabular}[c]{@{}c@{}}Holder \\ Table\end{tabular}} & $x_1$           & $\abs{8.05502}$   & 8.05508                & \textbf{8.05502}    & 8.05469    & \textbf{8.05502}    \\
                                                                         & $\sigma_{x_1}$  & NA                & 0.00015                & \textbf{0}          & 0.00067    & \textbf{0}          \\
                                                                         & $x_2$           & $\abs{9.66459}$   & 9.6646                 & \textbf{9.66459}    & 9.66453    & \textbf{9.66459}    \\
                                                                         & $\sigma_{x_2}$  & NA                & 0.00003                & \textbf{0}          & 0.00004    & \textbf{0}          \\
                                                                         & $f(x)$          & -19.208502        & -19.208502             & \textbf{-19.208502} & -19.208496 & \textbf{-19.208502} \\
                                                                         & $\sigma_{f(x)}$ & NA                & 0                      & \textbf{0}          & 0.000013   & \textbf{0}          \\ \midrule
\multirow{4}{*}{\begin{tabular}[c]{@{}c@{}}Cross In\\ Tray\end{tabular}} & $[x]$           & $\abs{[1.34940]}$ & 1.3494                 & \textbf{1.3494}     & 1.34924    & \textbf{1.3494}     \\
                                                                         & $\sigma_x$      & NA                & 0.00002                & \textbf{0}          & 0.00044    & \textbf{0}          \\
                                                                         & $f(x)$          & -2.062612         & -2.062612              & \textbf{-2.062612}  & -2.062612  & \textbf{-2.062612}  \\
                                                                         & $\sigma_{f(x)}$ & NA                & 0                      & \textbf{0}          & 0          & \textbf{0}          \\ \midrule
\multirow{4}{*}{\begin{tabular}[c]{@{}c@{}}Happy\\ Cat\end{tabular}}     & $x$             & {[}-1{]}          & -0.992                 & -0.996              & -0.993     & \textbf{-0.998}     \\
                                                                         & $\sigma_x$      & NA                & 0.127                  & 0.0916              & 0.116      & \textbf{0.0634}     \\
                                                                         & $f(x)$          & 0                 & 1.21E-02               & 0.0233              & 0.00252    & \textbf{'0.00106}   \\
                                                                         & $\sigma_{f(x)}$ & NA                & 0.0131                 & 0.00783             & 0.00529    & \textbf{0.00415}    \\ \midrule
\multirow{6}{*}{\begin{tabular}[c]{@{}c@{}}Schaffer \\ N4\end{tabular}}  & $x_1$           & 0                 & -0.113                 & \textbf{0}          & 0.00767    & \textbf{0}          \\
                                                                         & $\sigma_{x_1}$  & NA                & 0.38717                & \textbf{0}          & 0.02501    & \textbf{0}          \\
                                                                         & $x_2$           & 1.25312           & 1.30625                & \textbf{1.25313}    & 1.25333    & \textbf{1.25313}    \\
                                                                         & $\sigma_{x_2}$  & NA                & 0.05413                & \textbf{0}          & 0.00049    & \textbf{0}          \\
                                                                         & $f(x)$          & 0.292579          & 0.292693         & \textbf{0.292579}   & 0.292579   & \textbf{0.292579}   \\
                                                                         & $\sigma_{f(x)}$ & NA                & 0.000117               & \textbf{0}          & 0          & \textbf{0}          \\ \midrule
\multirow{4}{*}{6 Sphere}                                                & $[x]$           & {[}0{]}           & 2.00E-06               & \textbf{2.40E-166}  & 0.00122    & 1.0E-40             \\
                                                                         & $\sigma_x$      & NA                & 1.90E-05               & \textbf{0}          & 0.01548    & 1.70E-39            \\
                                                                         & $f(x)$          & 0                 & 2.1E-09                & \textbf{0}          & 0.00141    & 1.70E-77            \\
                                                                         & $\sigma_{f(x)}$ & NA                & 5.00E-09               & \textbf{0}          & 0.00053    & 4.20E-77            \\ \bottomrule
\end{tabular}
\caption{Each algorithm was run for 10 times per function and the solutions are read as mean $\pm$ standard deviation. The best results for each function are shown in bold.}
\label{tab:exacto}
\end{table*}

\begin{table}
\centering
\begin{tabular}{c|cccc}
              & GA & SOS & FA     & CS     \\ \hline
Schaffer N1   & 1  & 20  & $>$25  & $>$25  \\
Holder Table  & 1  & 8   & $>$12  & 3.5    \\
Cross-in-Tray & 1  & 20  & $>$30  & 23     \\
Schaffer N4   & 1  & 4   & $>$4.5 & $>$4.5 \\
6D Sphere     & 1  & 10  & $>$25  & $>$25 
\end{tabular}
\caption{ 
Relative time convergence of the algorithms, the normalization is made based on the fastest algorithms. The algorithms were run 10 times between 0.06 and 0.08 seconds, reporting the average value. Values with $>$ mean that the algorithm did not  converge in the given time.}
\label{tab:tiempo}
\end{table}

\section{Applications of AI modelling and metaheuristic optimization for NV drug delivery systems}

Of all the algorithms used to model and optimize NV properties for drug delivery systems, MLP-ANN and GA are respectively the most popular ones. Of the 16 articles mentioned in tables \ref{tab:resumen} and \ref{tab:resumen2}, 12 used MLP-ANN, 11 used GA and 7 used both. It is worth to note that, although not a majority, a significant number of papers did not mentioned parameters used for the MLP-ANN such as the number of hidden layers, the number of neurons of each layer and the activation function used for each layer. 

In 4 studies comparison was made between ANN and  RSM, in all of them MLP-ANN outperformed RSM, showing more accuracy in predicting the NV properties   \cite{Li2015, Varshosaz2012,  Saha2018, Zaki2015}. 
Comparison of different training algorithms for the MLP-ANN were also done. In one study, the authors compared the following training algorithms: Levenberg-Marquadt, Bayesian-Regularization, and Gradient Descent. ANNs trained with Bayesian-Regularization showed  slightly better predictive performance than the ones using Levenberg-Marquadt and far better results that the ones using gradient descent   \cite{Shahsavari2014}. In another study the following training algorithms of MLP-ANN were compared: quick propagation, Incremental Back Propagation, Batch Back Propagation, Levenberg-Marquadt, and GA. They concluded that GA was the most efficient one \cite{RoodbarShojaei2019}. Other metaheuristic algorithms that were used as a training algorithm for the AI algorithms were PSO \cite{Shafaei2020} and SA \cite{Hataminia2017}.

In 3 studies the optimization process was done experimentally using GA by an automated system, instead of modelling by AI or by polynomial regression methods. First, a series of experiments with initial conditions generated randomly were done and the properties were measured, then GA was used to generate the following set of conditions based on the conditions and properties of the last generation until a threshold value was met \cite{Salley2020, Lenders2017, Fernandes2016}.

Of the articles that applied modelling algorithms other than MLP-ANN, one article made two sets of experiments, the first one with a wide range of experimental conditions and was modelled with a fuzzy logic system in order to understand how the input variables and their interactions affected the NV properties, with this knowledge the search space was reduced and a second set of experiments was prepared, using MLP-ANN to model and GA to optimize \cite{Rouco2018}. Another article used SVM and GA as the modelling and optimization algorithms because common regression analyses and ANNs were not successful in predicting the properties of polymethacrylic acid-chitosan-polyethylene glycol NVs \cite{Rostamizadeh2015}.

\subsection{Testing modelling algorithms}
A comparison between ANFIS and MLP-ANN was made. ANFIS was chosen because it showed better performance than  MLP-ANN \cite{Ijadpanah-Saravi2017} and MLP-ANN was chosen because it is the most common machine learning algorithm used as shown in tables \ref{tab:resumen}, \ref{tab:resumen2} and \ref{tab:Comparacion}. The ANN was developed in MATLAB and the ANFIS was created with the MATLAB toolbox. These machine learning models where trained with an experimental data set of NP synthesis utilizing PLGA as the material for the NP, and Resveratrol as the encapsulating molecule (N=30) \cite{lozano2019optimization}. To define the structure of the MLP-ANN, its performance was evaluated for 2 to 5 neurons in the hidden layer and for two activation functions. The two activation functions compared are sigmoid eq.  \ref{eq:Act_sigmoid} and hyperbolic tangent eq. \ref{eq:Act_tangent} \cite{Glorot2011}. The performance of the algorithm was measured by 

\begin{equation}\label{eq:chi}
    \chi^2=\sum_{i=1}^N\left(\frac{(E_i-P_i)^2}{E_i}\right)
\end{equation}
where $E_i$ is the $i$th experimental data and $P_i$ is the $i$th predicted data. Four models were made to predict the following parameters: Diameter, PDI, ZP and Drug Loading Efficiency. Each network received three input variables: amount of PLGA [mg], PVA concentration and amount of resveratrol [mg]. The optimal activation function and number of hidden neurons for each network is shown in table \ref{tab:Opt_ANN}.

\begin{table}[ht]
\centering
\begin{tabular}{ccccc}
\hline
                                                                    & PS   & PDI  & PZ   & DLE  \\ \hline
\begin{tabular}[c]{@{}c@{}}Activation\\ function\end{tabular}       & tanh & tanh & tanh & tanh \\
\begin{tabular}[c]{@{}c@{}}Number of \\ hidden neurons\end{tabular} & 4    & 4    & 5    & 5    \\ \hline
\end{tabular}
\caption{Optimal parameters for ANN model.}
\label{tab:Opt_ANN}
\end{table}

The ANFIS parametrization was compared with the optimal formulation of the ANN, the results in table \ref{tab:chi} show that ANFIS has a better performance after training than the ANN model. These results are in agreement with those of table \ref{tab:Comparacion}. 

\begin{table}[H]
\centering
\begin{tabular}{@{}cllll@{}}
\toprule
$\chi^2$ & \multicolumn{1}{c}{PS} & \multicolumn{1}{c}{PDI} & \multicolumn{1}{c}{PZ} & \multicolumn{1}{c}{DLE} \\ \midrule
ANFIS    & 53.90            & 0.63              & 17.31             & 0.10              \\
ANN      & 57.23             & 0.68              & 17.66            & 0.11              \\ \bottomrule
\end{tabular}
\caption{Performance comparison between ANFIS and optimal ANN. The performance was measured with the $\chi^2$ parameter in eq. \ref{eq:chi}.}
\label{tab:chi}
\end{table}

\section{Methods to quantify the Sample Size Effects on performance of AI models}
The amount of data may be limited by resources availability or time to perform the experiments. Because of this, studies with low sample size data may be questioned \cite{balki2019sample}. However, there is the contrary case when there is a huge data set and, in this case, limiting the amount of data that is used might be necessary to avoid overfitting. This section will provide an overview of methodologies that can be applied to evaluate the accuracy of a given modelling method.

\subsection{Sub-sampling}
\begin{figure}[ht]
    \centering
    \includegraphics[scale=0.45]{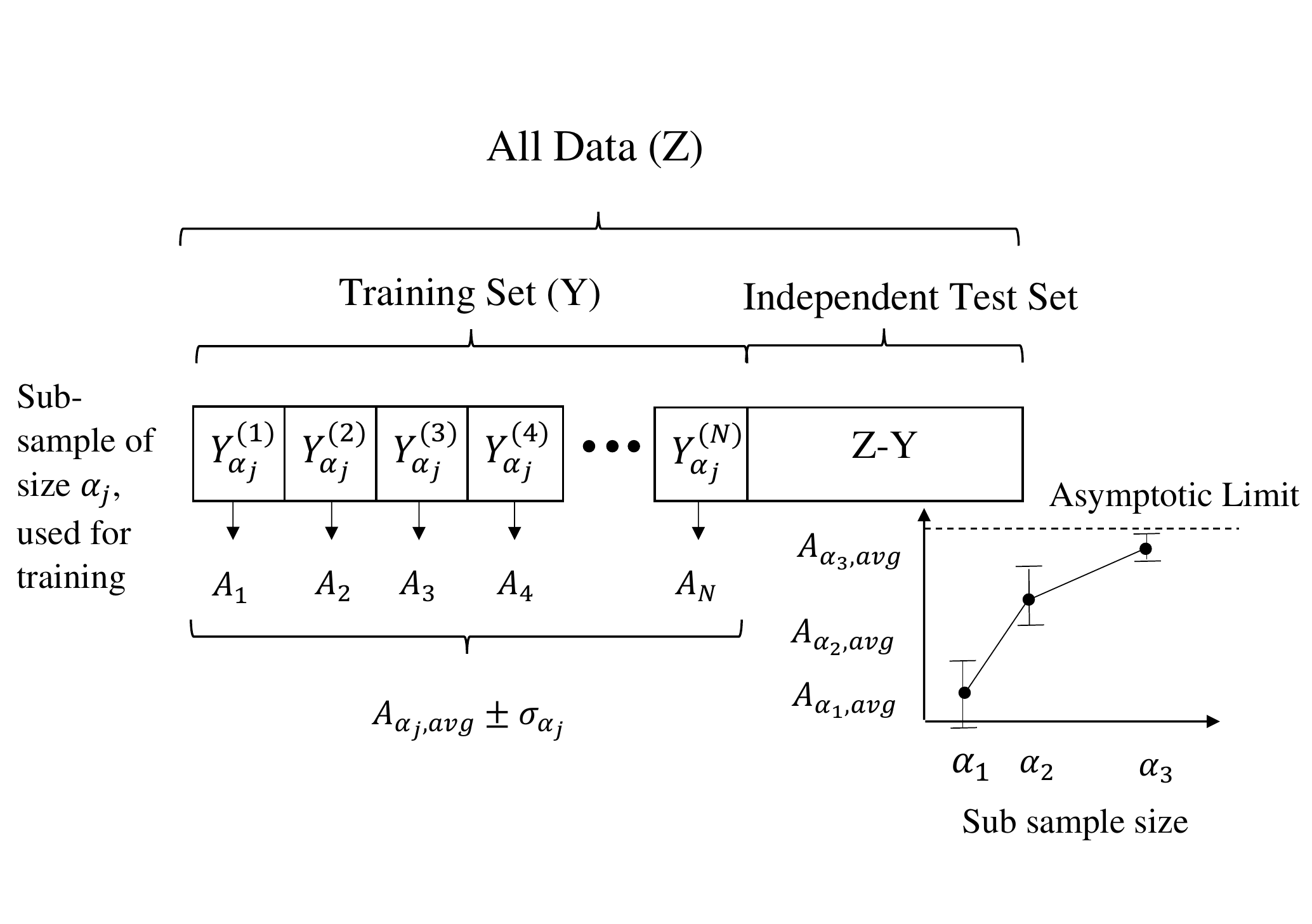}
    \caption{Scheme of Sub-sampling.}
    \label{fig:subsampling}
\end{figure}

This method consists in splitting the whole data, of size Z, into a the training data and the test data, of size \textit{Y} and \textit{Z}-\textit{Y}, respectively. After that, the training data is divided again in subsets of equal size. Then, each subset is evaluated with the test data. The process is repeated for a new size of subsets. Accuracy and standard deviation with respect to the test data is measured in every cycle. A scheme of this method is shown in figure \ref{fig:subsampling}. This method has been applied in studies involving machine learning such as energy consumption in buildings modelling \cite{macas2016role} and for reconstruction of multi-dimensional magnetic resonance fingerprinting \cite{cohen2018mr}.

\subsection{Repeated Cross-Validation}
\begin{figure}[ht]
    \centering
    \includegraphics[scale=0.45]{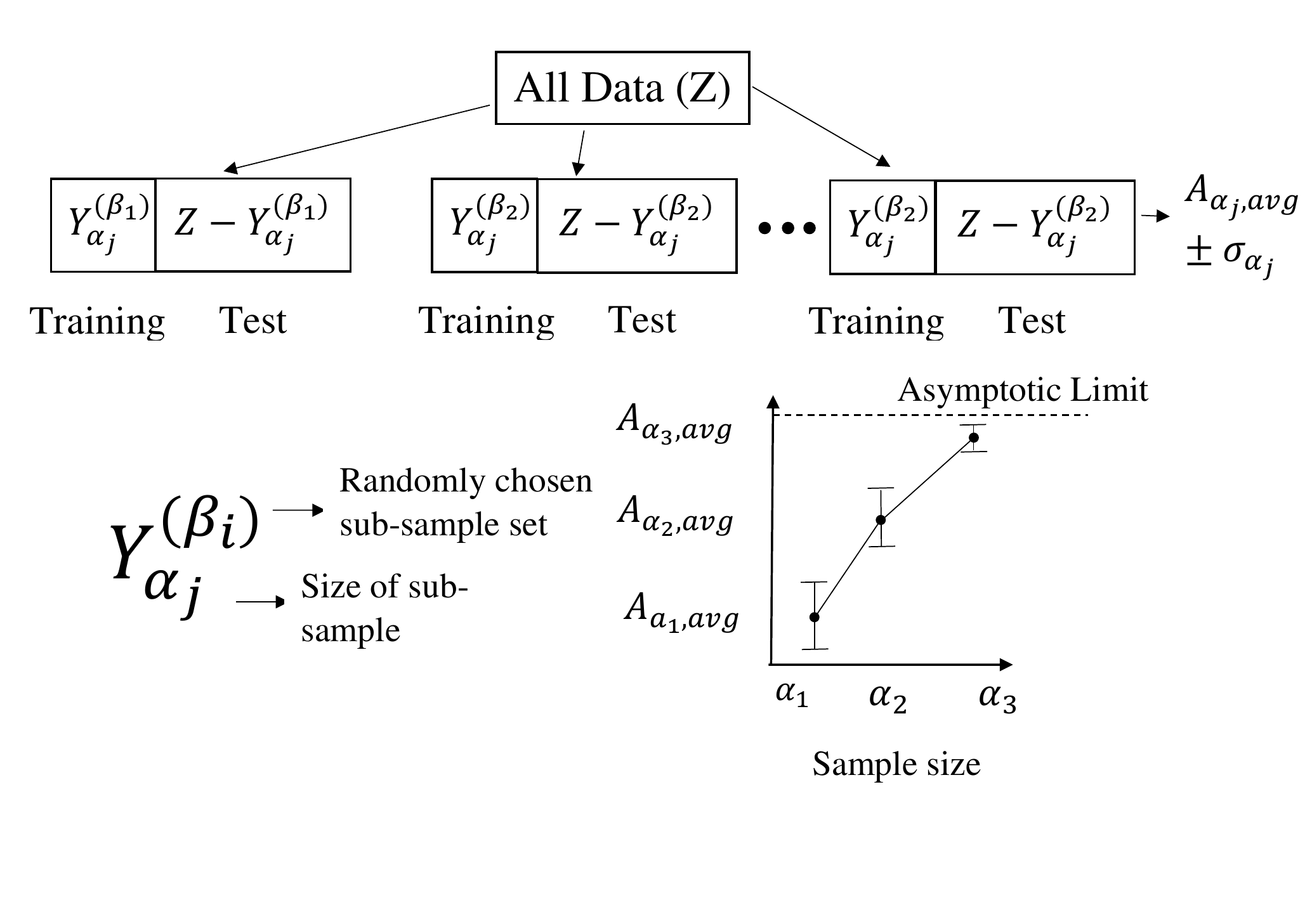}
    \caption{Scheme of Cross validation method.}
    \label{fig:cross_validationl}
\end{figure}

In this method the data of size \textit{Z}, is splitted randomly into a training and test set, of size \textit{Y$_{\alpha_j}$} and \textit{Z-Y$_{\alpha_j}$} respectively. This is done N times in order to obtain a standard deviation for each Y$_{\alpha_j}$. This is repeated k times with a different training size \textit{Y$_j$} where j=1,2,3,...,k and the average performance is compared \cite{balki2019sample,tourassi1997effect,kohavi1995study}. A visual representation is shown in figure \ref{fig:cross_validationl}. 

\subsection{No repetition scheme}
\begin{figure}[H]
    \centering
    \includegraphics[scale=0.45]{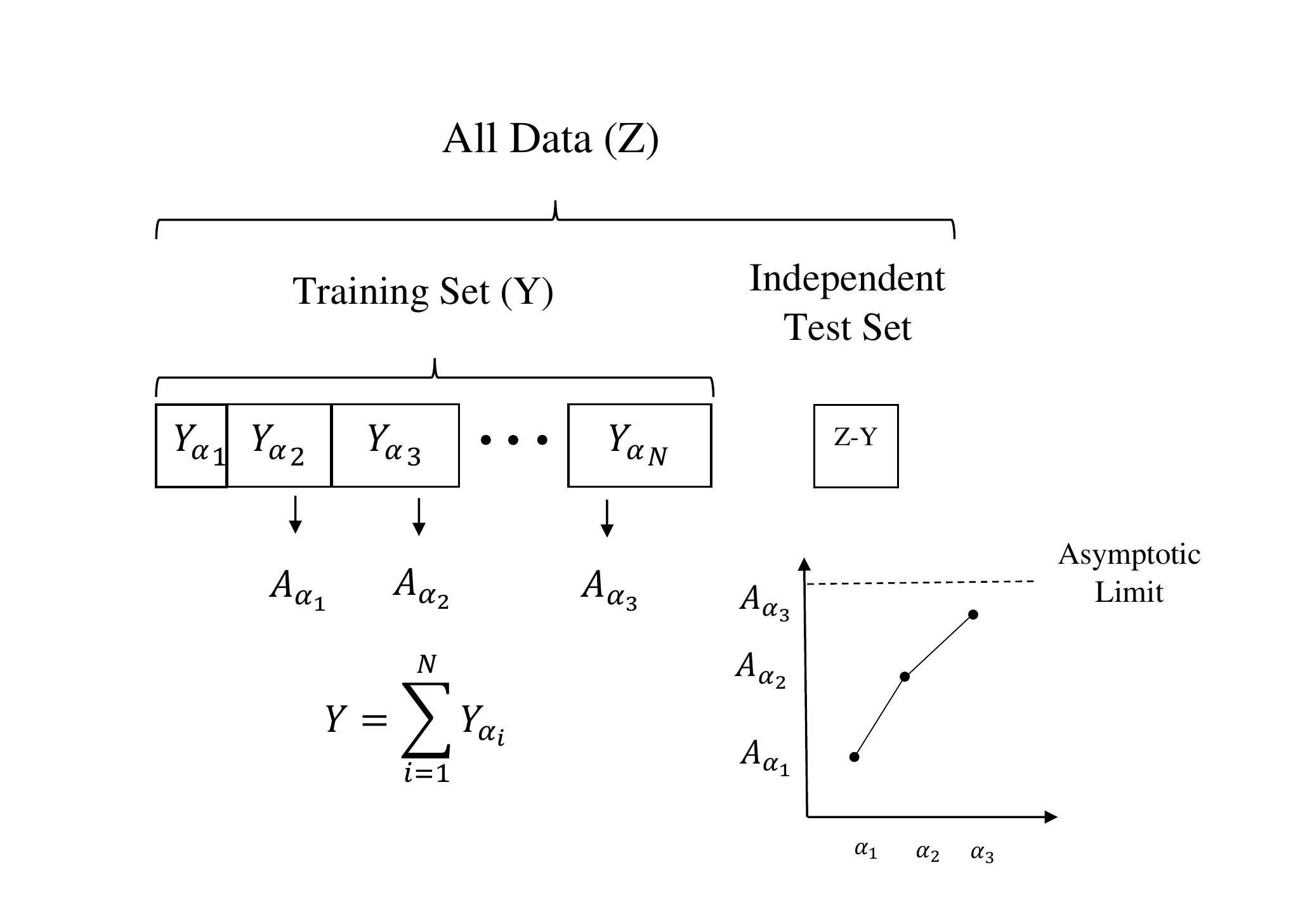}
    \caption{Scheme for no repetition method.}
    \label{fig:no_repetition}
\end{figure}

This method is similar to the \textit{Sub-Sampling} method. The data of size Z, is divided into a training and test set, of sizes Y and Y-Z respectively. The training set is divided again in different sizes $Y_{\alpha_j}$. The accuracy is measured for each training size. Figure \ref{fig:no_repetition} gives a visual representation of this method. No repetition scheme have been used to analyze the sample effects on handwriting character recognition \cite{liu2002performance}  to obtain high-resolution pixel-wise brain segmentation \cite{zhao2018bayesian}, for species distributions models \cite{wisz2008effects}, among other applications.   

\subsection{Critical Sampling Size (CSS) Heuristic Methods}
This method is based on the existence of a large data pool, from which a subset of data will be selected for analysis, avoiding redundant data \cite{silva2017finding}. The CSS heuristic method consists on dividing the data sample of size \textit{Z} into \textit{k} clusters, taking \textit{m} data-points randomly from each cluster. \textit{k} could be chosen to be the number of classes of the data. This process will form a training data set of size \textit{m} $\times$ \textit{k}$=D_{T_m}$. Furthermore, a \textit{d} $\times$ \textit{k}$=D_{T_d}$ set of data is added to the training set. This last step is to provide a lower limit to the training sample size. If the performance of the trained model achieves a predefined threshold value T (for example 2\% \cite{alwosheel2018your}) on an independent data set, then the training data size is considered to be big enough. Otherwise, this process is repeated with larger values of m, until the model satisfies the desired accuracy. A more detailed description of CSS heuristic method can be found at \cite{sung2016sampling}.

\begin{figure}[H]
    \centering
    \includegraphics[scale=0.45]{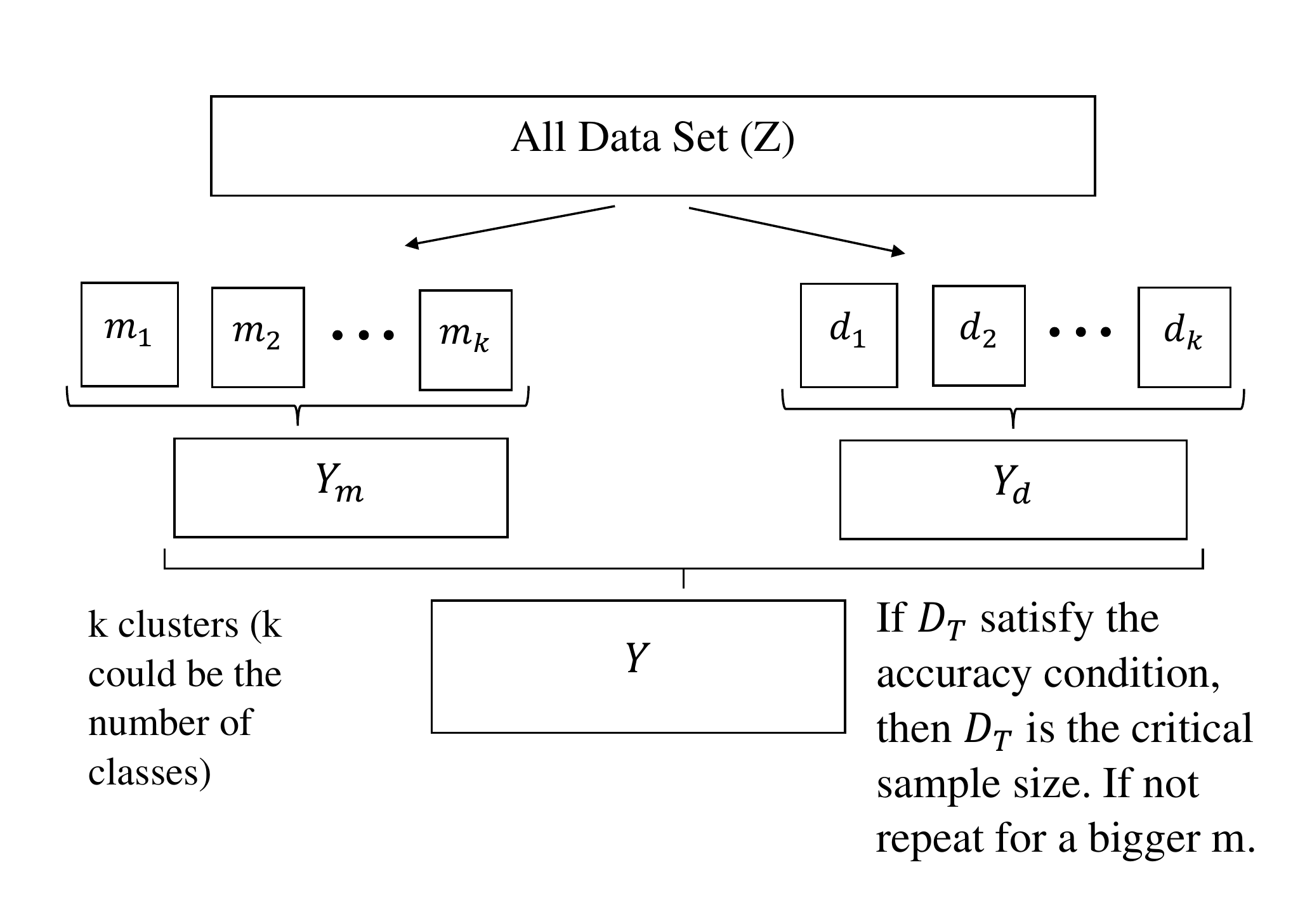}
    \caption{Scheme of the critical sampling size heuristic method}
    \label{fig:my_label}
\end{figure}

\section{Conclusions}

Finding the most optimal conditions for a NV could represent the difference between success and failure in research and development. In this regard, the use of AI and metaheuristic algorithms for NVs drug delivery systems design is a relative new field. This is reflected through several areas of opportunity for improvement. Firstly, is the use of relatively popular but not the most efficient algorithms, such as MLP-ANN or GA, which is directly related to the low amount of algorithm comparison for NV design, especially metaheuristic algorithms. As shown here, for AI algorithms ANFIS could be better than MLP-ANN, in agreement with a previous study; and for metaheuristic algorithms, SOS and CS clearly outperform GA. Secondly, all information from a specific algorithm should be provided, for example in MLP-ANN the parameters of activation function, number of hidden layers, input and output variables should be included, or if a commercial package is used then details on the brand and model should be made available. Finally, sample size appropriateness assessment would provide an indication of the relative confidence on the NV modelling from an specific AI algorithm.

\section{Acknowledgments}
Roberto Quintanilla. CONACYT project, premio tec salud project

\bibliographystyle{naturemag}
\bibliography{main}

\end{document}